\documentstyle[aclap,epic]{article}
\catcode`\"=\active \let"=\" \let\3=\ss

\title{\vspace{-0.5in}Identifying Word Translations in Non-Parallel Texts}
\author{Reinhard Rapp \\
ISSCO, Universit\'e de Gen\`{e}ve\\
54 route des Acacias\\
Gen\`{e}ve, Switzerland\\
rapp@divsun.unige.ch\\}

\begin{document}

\maketitle
\vspace{-0.5in}
\begin{abstract}

Common algorithms for sentence and word-alignment allow the automatic
identification
of word translations from parallel texts. This study suggests that the
identification
of word translations should also be possible with non-parallel and even
unrelated texts. The
method proposed is based on the assumption that there is a correlation between
the patterns
of word co-occurrences in texts of different languages.
\end{abstract}

\section{Introduction}
In a number of recent studies it has been shown that word translations can be
automatically
derived from the statistical distribution of words in bilingual parallel texts
(e.~g.~Catizone,
Russell \& Warwick, 1989; Brown et al., 1990; Dagan, Church \& Gale, 1993; Kay
\& R"oscheisen, 1993).
Most of the proposed algorithms first conduct an alignment of sentences,
i.~e.~those pairs of sentences
are located that are translations of each other. In a second step a word
alignment is performed
by analyzing the correspondences of words in each pair of sentences.

The results achieved with these algorithms have been found useful for the
compilation of
dictionaries, for checking the consistency of terminological usage in
translations, and for
assisting the terminological work of translators and interpreters.

However, despite serious efforts in the compilation of corpora (Church \&
Mercer, 1993;
Armstrong \& Thompson, 1995) the availability of a large enough parallel corpus
in a specific field and for a given pair of languages will always be the
exception, not the rule.
Since the acquisition of non-parallel texts is usually much easier, it would be
desirable to have a
program that can determine the translations of words from comparable or even
unrelated texts.

\vspace{0.3cm}

\section{Approach}

It is assumed that there is a correlation between the co-occurrences of words
which are
translations of each other. If -- for example -- in a text of one language two
words {\em A} and
{\em B} co-occur more often than expected from chance, then in a text of
another language
those words which are translations of {\em A} and {\em B} should also co-occur
more
frequently than expected. This assumption is reasonable for parallel texts.
However, in this
paper it is further assumed that the co-occurrence patterns in original texts
are not
fundamentally different from those in translated texts.

Starting from an English vocabulary of six words and the corresponding German
translations,
table~\ref{matrixpermutation}a and b show
an English and a German co-occurrence matrix. In these matrices the
entries belonging to those pairs of words that in texts co-occur more
frequently than expected
have been marked with a dot. In general, word order in the lines and columns of
a
co-occurrence matrix is independent of each other, but for the purpose of this
paper
can always be assumed to be equal without loss of generality.

\begin{table}
\begin{center}
\vspace{-0.43cm}
\caption{When the word orders of the English and the German matrix correspond,
the dot patterns of the two matrices are identical.}

\vspace{0.1cm}
\label{matrixpermutation}
\parbox{0.8cm}{(a)}\parbox{8.3cm}{\begin{tabular}{|lr||c|c|c|c|c|c|} \hline
\hspace{0.31cm} & & 1 &  2    &   3   &   4  & 5      & 6      \\ \hline \hline
blue    & 1 & \hspace{0.31cm} &$\bullet$&  &         &$\bullet$&          \\
\hline
green   & 2 &$\bullet$& \hspace{0.31cm} &$\bullet$ & &         &          \\
\hline
plant   & 3 &         &$\bullet$& \hspace{0.31cm} &  &         &          \\
\hline
school  & 4 &         &         &         & \hspace{0.31cm} &  &$\bullet$ \\
\hline
sky     & 5 &$\bullet$&         &         &         & \hspace{0.31cm} &   \\
\hline
teacher & 6 &         &         &         &$\bullet$&  &\hspace{0.31cm}   \\
\hline
\end{tabular}}

\vspace{0.5cm}

\parbox{0.8cm}{(b)}\parbox{13.3cm}{\begin{tabular}{|lr||c|c|c|c|c|c|} \hline
\hspace{0.31cm} & & 1 &  2    &   3   &   4  & 5      & 6      \\ \hline \hline
blau    & 1 & \hspace{0.31cm} &$\bullet$&$\bullet$&  &         &          \\
\hline
gr"un   & 2 &$\bullet$& \hspace{0.31cm} &  &         &$\bullet$&          \\
\hline
Himmel \hspace*{-0.2cm}  & 3 &$\bullet$&         & \hspace{0.31cm} &  &
&          \\ \hline
Lehrer  & 4 &         &         &         & \hspace{0.31cm} & &$\bullet$  \\
\hline
Pflanze & 5 &         &$\bullet$&         &  & \hspace{0.31cm} &          \\
\hline
Schule  & 6 &         &         &         &$\bullet$& & \hspace{0.31cm}   \\
\hline
\end{tabular}}

\vspace{0.5cm}

\parbox{0.8cm}{(c)}\parbox{13.3cm}{\begin{tabular}{|lr||c|c|c|c|c|c|r} \hline
\hspace{0.31cm} & & 1 &  2    &   5   &   6  & 3      & 4      \\ \hline \hline
blue    & 1 & \hspace{0.31cm} &$\bullet$&$\bullet$&  &         &          \\
\hline
green   & 2 &$\bullet$& \hspace{0.31cm} &  &         &$\bullet$&          \\
\hline
sky     & 5 &$\bullet$&         & \hspace{0.31cm} &  &         &          \\
\hline
teacher & 6 &         &         &         & \hspace{0.31cm} & &$\bullet$  \\
\hline
plant   & 3 &         &$\bullet$&         &  & \hspace{0.31cm} &          \\
\hline
school  & 4 &         &         &         &$\bullet$& & \hspace{0.31cm}   \\
\hline
\end{tabular}}

\vspace{-0.4cm}
\end{center}
\end{table}

If now the word order of the English matrix is permuted until the resulting
pattern of dots is
most similar to that of the German matrix (see table~\ref{matrixpermutation}c),
then this increases the likelihood
that the English and German words are in corresponding order. Word {\em n} in
the English
matrix is then the translation of word {\em n} in the German matrix.

\section{Simulation}

A simulation experiment was conducted in order to see whether the above
assumptions
concerning the similarity of co-occurrence patterns actually hold. In this
experiment, for
an equivalent English and German vocabulary two co-occurrence matrices were
computed and
then compared. As the English vocabulary a list of 100 words was used, which
had been
suggested by Kent \& Rosanoff (1910) for association experiments. The German
vocabulary consisted of one by one translations of these words as chosen by
Russell (1970).

The word co-occurrences were computed on the basis of an English corpus of 33
and a
German corpus of 46 million words. The English corpus consists of the {\em
Brown Corpus}, texts
from the {\em Wall Street Journal}, {\em Grolier's Electronic Encyclopedia} and
scientific abstracts from
different fields. The German corpus is a compilation of mainly newspaper texts
from {\em
Frankfurter Rundschau}, {\em Die Zeit} and {\em Mannheimer Morgen}. To the
knowledge of the
author, the English and German corpora contain no parallel passages.

For each pair of words in the English vocabulary its frequency of common
occurrence in the
English corpus was counted. The common occurrence of two words was defined as
both words
being separated by at most 11 other words. The co-occurrence frequencies
obtained in this way
were used to build up the English matrix. Equivalently, the German
co-occurrence matrix was
created by counting the co-occurrences of German word pairs in the German
corpus. As a
starting point, word order in the two matrices was chosen such that word {\em
n} in
the German matrix was the translation of word {\em n} in the English matrix.

Co-occurrence studies like that conducted by Wett\-ler \& Rapp (1993) have
shown that for many
purposes it is desirable to reduce the influence of word frequency on the
co-occurrence counts.
For the prediction of word associations they achieved best results when
modifying each
entry in the co-occurrence matrix using the following formula:
\begin{equation}
A_{i,j} \ = \ \frac{(f(i \& j))^2}{f(i) \cdot f(j)}
\label{formula1}
\end{equation}
Hereby $f(i \& j)$ is the frequency of common occurrence of the two words $i$
and $j$, and $f(i)$ is
the corpus frequency of word $i$. However, for comparison, the simulations
described below
were also conducted using the original co-occurrence matrices
(formula~\ref{formula2}) and
a measure similar to mutual information (formula~\ref{formula3}).\footnote{The
logarithm has
been removed from the mutual information measure since it is not defined for
zero co-occurrences.}
\begin{equation}
A_{i,j} \ = \ f(i \& j)
\label{formula2}
\end{equation}

\begin{equation}
A_{i,j} \ = \ \frac{f(i \& j)}{f(i) \cdot f(j)}
\label{formula3}
\end{equation}
Regardless of the formula applied, the English and the German matrix were both
normalized.\footnote{Normalization was conducted in such a way that the sum of
all matrix
entries adds up to the number of fields in the matrix.} Starting from the
normalized English and
German matrices, the aim was to determine how far the similarity of the two
matrices
depends on the correspondence of word order. As a measure for matrix similarity
the sum of
the absolute differences of the values at corresponding matrix positions was
used.
\begin{equation}
s = \sum^N_{i=1} \sum^N_{j=1} |E_{i,j}-G_{i,j}|
\end{equation}
This similarity measure leads to a value of zero for identical matrices, and to
a value of 20 000
in the case that a non-zero entry in one of the 100 * 100 matrices always
corresponds to a
zero-value in the other.

\section{Results}

\begin{figure}
\begin{center}
\unitlength0.5mm
\begin{picture}(155.000,100.000)
\put(0,0){\framebox(155.000, 100.000)}
\put(20.000,10.000){\vector(1,0){125.000}}
\put(147.000,10.000){\makebox(0,0)[l]{$c$}}
\put(25.000,10.000){\line(0,-1){1.000}}
\put(25.000,7.000){\makebox(0,0)[t]{0}}
\put(36.500,10.000){\line(0,-1){1.000}}
\put(36.500,7.000){\makebox(0,0)[t]{10}}
\put(48.000,10.000){\line(0,-1){1.000}}
\put(48.000,7.000){\makebox(0,0)[t]{20}}
\put(59.500,10.000){\line(0,-1){1.000}}
\put(59.500,7.000){\makebox(0,0)[t]{30}}
\put(71.000,10.000){\line(0,-1){1.000}}
\put(71.000,7.000){\makebox(0,0)[t]{40}}
\put(82.500,10.000){\line(0,-1){1.000}}
\put(82.500,7.000){\makebox(0,0)[t]{50}}
\put(94.000,10.000){\line(0,-1){1.000}}
\put(94.000,7.000){\makebox(0,0)[t]{60}}
\put(105.500,10.000){\line(0,-1){1.000}}
\put(105.500,7.000){\makebox(0,0)[t]{70}}
\put(117.000,10.000){\line(0,-1){1.000}}
\put(117.000,7.000){\makebox(0,0)[t]{80}}
\put(128.500,10.000){\line(0,-1){1.000}}
\put(128.500,7.000){\makebox(0,0)[t]{90}}
\put(140.000,10.000){\line(0,-1){1.000}}
\put(140.000,7.000){\makebox(0,0)[t]{100}}
\put(20.000,10.000){\vector(0,1){80.000}}
\put(20.000,92.000){\makebox(0,0)[b]{$\bar{s} / 1000$ \ }}
\put(20.000,15.000){\line(-1,0){1.000}}
\put(17.000,15.000){\makebox(0,0)[r]{10}}
\put(20.000,22.000){\line(-1,0){1.000}}
\put(17.000,22.000){\makebox(0,0)[r]{}}
\put(20.000,29.000){\line(-1,0){1.000}}
\put(17.000,29.000){\makebox(0,0)[r]{12}}
\put(20.000,36.000){\line(-1,0){1.000}}
\put(17.000,36.000){\makebox(0,0)[r]{}}
\put(20.000,43.000){\line(-1,0){1.000}}
\put(17.000,43.000){\makebox(0,0)[r]{14}}
\put(20.000,50.000){\line(-1,0){1.000}}
\put(17.000,50.000){\makebox(0,0)[r]{}}
\put(20.000,57.000){\line(-1,0){1.000}}
\put(17.000,57.000){\makebox(0,0)[r]{16}}
\put(20.000,64.000){\line(-1,0){1.000}}
\put(17.000,64.000){\makebox(0,0)[r]{}}
\put(20.000,71.000){\line(-1,0){1.000}}
\put(17.000,71.000){\makebox(0,0)[r]{18}}
\put(20.000,78.000){\line(-1,0){1.000}}
\put(17.000,78.000){\makebox(0,0)[r]{}}
\put(20.000,85.000){\line(-1,0){1.000}}
\put(17.000,85.000){\makebox(0,0)[r]{20}}

\small
\put(147.0,67.70){\makebox(0,0)[c]{(2)}}
\put(147.0,75.60){\makebox(0,0)[c]{(3)}}
\put(147.0,83.50){\makebox(0,0)[c]{(1)}}
\normalsize

\drawline(25.000,37.493)(27.300,38.465)(28.450,38.858)
(29.600,39.387)(30.750,39.894)(31.900,40.299)(33.050,40.820)
(34.200,41.073)(35.350,41.639)(36.500,41.987)(37.650,42.441)
(38.800,42.962)(39.950,43.395)(41.100,43.715)(42.250,44.205)
(43.400,44.632)(44.550,45.079)(45.700,45.438)(46.850,45.866)
(48.000,46.354)(49.150,46.799)(50.300,47.181)(51.450,47.676)
(52.600,47.938)(53.750,48.471)(54.900,48.848)(56.050,49.222)
(57.200,49.541)(58.350,50.135)(59.500,50.240)(60.650,50.863)
(61.800,51.196)(62.950,51.621)(64.100,51.940)(65.250,52.275)
(66.400,52.737)(67.550,53.096)(68.700,53.350)(69.850,53.742)
(71.000,54.097)(72.150,54.453)(73.300,54.784)(74.450,55.113)
(75.600,55.397)(76.750,55.639)(77.900,55.906)(79.050,56.557)
(80.200,56.533)(81.350,57.066)(82.500,57.391)(83.650,57.634)
(84.800,57.885)(85.950,58.365)(87.100,58.682)(88.250,58.768)
(89.400,59.248)(90.550,59.410)(91.700,59.552)(92.850,59.929)
(94.000,60.323)(95.150,60.689)(96.300,60.912)(97.450,61.331)
(98.600,61.566)(99.750,61.805)(100.900,62.049)(102.050,62.666)
(103.200,62.485)(104.350,62.768)(105.500,63.014)(106.650,63.305)
(107.800,63.525)(108.950,63.725)(110.100,63.917)(111.250,64.345)
(112.400,64.502)(113.550,64.979)(114.700,65.238)(115.850,65.060)
(117.000,65.489)(118.150,65.705)(119.300,66.043)(120.450,66.039)
(121.600,66.260)(122.750,66.390)(123.900,66.585)(125.050,67.036)
(126.200,66.977)(127.350,66.946)(128.500,67.357)(129.650,67.353)
(130.800,67.263)(131.950,67.419)(133.100,67.722)(134.250,68.165)
(135.400,68.663)(136.550,68.712)(137.700,68.451)(138.850,68.604)
(140.000,68.780)

\drawline(25.000,42.887)(27.300,44.021)(28.450,44.616)
(29.600,45.196)(30.750,45.786)(31.900,46.411)(33.050,46.837)
(34.200,47.594)(35.350,47.950)(36.500,48.759)(37.650,48.957)
(38.800,49.726)(39.950,50.174)(41.100,50.834)(42.250,51.029)
(43.400,51.951)(44.550,52.111)(45.700,52.711)(46.850,53.353)
(48.000,53.686)(49.150,54.314)(50.300,54.706)(51.450,55.238)
(52.600,55.745)(53.750,55.987)(54.900,56.534)(56.050,56.992)
(57.200,57.291)(58.350,58.002)(59.500,58.169)(60.650,58.761)
(61.800,59.153)(62.950,59.413)(64.100,59.968)(65.250,60.258)
(66.400,60.802)(67.550,60.937)(68.700,61.609)(69.850,62.187)
(71.000,62.314)(72.150,62.874)(73.300,63.001)(74.450,63.319)
(75.600,63.876)(76.750,64.149)(77.900,64.579)(79.050,64.583)
(80.200,65.193)(81.350,65.373)(82.500,65.822)(83.650,66.063)
(84.800,66.271)(85.950,66.394)(87.100,67.054)(88.250,66.977)
(89.400,67.536)(90.550,67.798)(91.700,68.156)(92.850,68.355)
(94.000,68.727)(95.150,69.053)(96.300,69.009)(97.450,69.327)
(98.600,69.726)(99.750,70.006)(100.900,69.817)(102.050,70.623)
(103.200,70.664)(104.350,70.915)(105.500,70.818)(106.650,71.514)
(107.800,71.409)(108.950,71.498)(110.100,71.805)(111.250,71.936)
(112.400,72.154)(113.550,72.127)(114.700,72.636)(115.850,72.446)
(117.000,72.598)(118.150,72.827)(119.300,72.875)(120.450,72.893)
(121.600,72.973)(122.750,73.143)(123.900,73.438)(125.050,73.327)
(126.200,73.574)(127.350,73.699)(128.500,73.625)(129.650,73.787)
(130.800,73.895)(131.950,73.995)(133.100,73.994)(134.250,73.959)
(135.400,74.111)(136.550,74.435)(137.700,74.309)(138.850,74.520)

\drawline(25.000,21.356)(27.300,23.530)(28.450,24.531)
(29.600,26.167)(30.750,26.905)(31.900,28.679)(33.050,29.283)
(34.200,30.920)(35.350,31.691)(36.500,33.274)(37.650,33.661)
(38.800,35.019)(39.950,35.990)(41.100,37.195)(42.250,37.397)
(43.400,39.056)(44.550,39.712)(45.700,40.866)(46.850,41.856)
(48.000,42.676)(49.150,43.829)(50.300,44.884)(51.450,46.039)
(52.600,46.574)(53.750,47.974)(54.900,48.387)(56.050,49.427)
(57.200,50.205)(58.350,51.140)(59.500,52.017)(60.650,52.823)
(61.800,53.763)(62.950,54.377)(64.100,55.810)(65.250,56.192)
(66.400,57.075)(67.550,57.444)(68.700,58.720)(69.850,59.861)
(71.000,59.736)(72.150,61.284)(73.300,61.225)(74.450,62.047)
(75.600,63.149)(76.750,63.494)(77.900,64.097)(79.050,64.356)
(80.200,65.309)(81.350,65.972)(82.500,66.483)(83.650,67.416)
(84.800,67.420)(85.950,68.195)(87.100,69.538)(88.250,69.007)
(89.400,70.488)(90.550,70.523)(91.700,71.066)(92.850,71.572)
(94.000,72.330)(95.150,72.422)(96.300,73.029)(97.450,73.625)
(98.600,74.339)(99.750,74.712)(100.900,74.731)(102.050,75.703)
(103.200,75.609)(104.350,75.937)(105.500,76.242)(106.650,77.136)
(107.800,77.165)(108.950,77.386)(110.100,78.082)(111.250,78.348)
(112.400,78.591)(113.550,78.787)(114.700,79.294)(115.850,79.325)
(117.000,79.749)(118.150,79.800)(119.300,80.025)(120.450,80.100)
(121.600,80.308)(122.750,80.434)(123.900,81.060)(125.050,81.244)
(126.200,80.851)(127.350,81.418)(128.500,81.465)(129.650,81.759)
(130.800,81.709)(131.950,81.805)(133.100,81.873)(134.250,82.114)
(135.400,82.184)(136.550,82.311)(137.700,82.311)(138.850,82.393)
(140.000,82.304)

\dottedline(25.000,21.356)(27.300,20.601)(28.450,21.245)
(29.600,18.898)(30.750,19.474)(31.900,19.175)(33.050,21.236)
(34.200,19.368)(35.350,19.349)(36.500,19.322)(37.650,19.800)
(38.800,19.345)(39.950,21.690)(41.100,19.376)(42.250,18.243)
(43.400,19.500)(44.550,19.004)(45.700,18.942)(46.850,22.295)
(48.000,18.973)(49.150,21.679)(50.300,19.100)(51.450,22.891)
(52.600,19.329)(53.750,19.390)(54.900,23.391)(56.050,19.524)
(57.200,23.215)(58.350,19.574)(59.500,23.440)(60.650,23.714)
(61.800,23.795)(62.950,23.718)(64.100,23.688)(65.250,24.461)
(66.400,24.507)(67.550,26.557)(68.700,25.861)(69.850,26.275)
(71.000,25.947)(72.150,26.106)(73.300,31.199)(74.450,26.272)
(75.600,30.472)(76.750,26.730)(77.900,30.170)(79.050,30.181)
(80.200,30.185)(81.350,30.199)(82.500,30.223)(83.650,33.745)
(84.800,31.041)(85.950,37.840)(87.100,33.909)(88.250,38.086)
(89.400,33.839)(90.550,39.407)(91.700,33.903)(92.850,39.967)
(94.000,41.409)(95.150,40.278)(96.300,40.273)(97.450,40.361)
(98.600,40.368)(99.750,40.421)(100.900,40.435)(102.050,47.176)
(103.200,47.147)(104.350,46.924)(105.500,47.255)(106.650,47.254)
(107.800,47.252)(108.950,47.463)(110.100,49.615)(111.250,50.167)
(112.400,50.168)(113.550,50.021)(114.700,50.010)(115.850,50.298)
(117.000,58.290)(118.150,58.278)(119.300,62.826)(120.450,64.205)
(121.600,63.832)(122.750,63.787)(123.900,63.756)(125.050,63.920)
(126.200,63.966)(127.350,64.475)(128.500,71.162)(129.650,70.961)
(130.800,70.518)(131.950,69.567)(133.100,70.410)(134.250,70.405)
(135.400,70.436)(136.550,71.006)(137.700,70.986)(138.850,78.961)
(140.000,75.606)

\dottedline(25.000,21.356)(27.300,49.931)(28.450,50.012)
(29.600,53.586)(30.750,53.062)(31.900,57.683)(33.050,59.950)
(34.200,61.715)(35.350,62.936)(36.500,65.394)(37.650,65.509)
(38.800,63.970)(39.950,66.064)(41.100,71.057)(42.250,71.204)
(43.400,69.815)(44.550,71.228)(45.700,69.969)(46.850,73.145)
(48.000,70.621)(49.150,73.318)(50.300,72.099)(51.450,73.358)
(52.600,73.415)(53.750,75.278)(54.900,75.450)(56.050,75.444)
(57.200,74.516)(58.350,79.161)(59.500,76.498)(60.650,79.295)
(61.800,76.895)(62.950,79.300)(64.100,79.284)(65.250,79.358)
(66.400,79.582)(67.550,79.553)(68.700,79.804)(69.850,79.829)
(71.000,79.806)(72.150,80.063)(73.300,80.789)(74.450,80.811)
(75.600,80.373)(76.750,80.963)(77.900,81.010)(79.050,80.514)
(80.200,81.040)(81.350,81.260)(82.500,81.408)(83.650,81.905)
(84.800,81.540)(85.950,82.007)(87.100,82.030)(88.250,82.018)
(89.400,82.093)(90.550,82.163)(91.700,82.204)(92.850,82.266)
(94.000,82.235)(95.150,82.285)(96.300,82.309)(97.450,82.472)
(98.600,82.470)(99.750,82.431)(100.900,82.472)(102.050,82.502)
(103.200,82.534)(104.350,82.507)(105.500,82.429)(106.650,82.606)
(107.800,82.676)(108.950,82.981)(110.100,82.808)(111.250,82.978)
(112.400,82.986)(113.550,83.009)(114.700,83.013)(115.850,83.072)
(117.000,83.092)(118.150,83.169)(119.300,83.179)(120.450,83.172)
(121.600,83.177)(122.750,83.180)(123.900,83.131)(125.050,83.064)
(126.200,83.127)(127.350,83.116)(128.500,83.115)(129.650,83.123)
(130.800,83.130)(131.950,83.148)(133.100,83.096)(134.250,83.304)
(135.400,83.252)(136.550,83.228)(137.700,83.283)(138.850,83.235)
(140.000,83.359)

\end{picture}
\end{center}
\vspace{-0.3cm}
\caption{Dependency between the mean similarity~$\bar{s}$\linebreak
of the English
and the German matrix and the num\-ber of non-corresponding word
positions $c$ for 3~for\-mulas. The dotted lines are the minimum
and maximum values of each sample of 1000 for formula 1.}
\label{curves}
\end{figure}
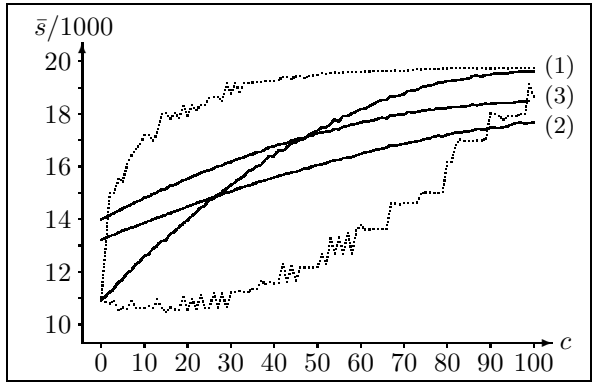

The simulation was conducted by randomly permuting the word order of the German
matrix
and then computing the similarity {\em s} to the English matrix. For each
permutation it was
determined how many words {\em c} had been shifted to positions different from
those in the
original German matrix. The simulation was continued until for each value of
{\em c} a set of
1000 similarity values was available.\footnote{$c = 1$ is not possible and was
not taken into account.} Figure~\ref{curves} shows for the three
formulas how the average similarity $\bar{s}$ between the English and the
German matrix
depends on the number of non-corresponding word positions {\em c}. Each of the
curves increases
monotonically, with formula~\ref{formula1} having the steepest, i.~e.~best
discriminating characteristic.
The dotted curves in figure~\ref{curves} are the minimum and maximum values in
each set of 1000
similarity values for formula~\ref{formula1}.

\section{Discussion and prospects}

It could be shown that even for unrelated Eng\-lish and German texts the
patterns
of word co-occurrences strongly correlate. The monotonically increasing
character
of the curves in figure~\ref{curves} indicates that in principle it should be
possible to find word correspondences in two matrices of different languages by
randomly permuting one of the matrices until the similarity function $s$
reaches
a minimum and thus indicates maximum similarity.
However, the minimum-curve in figure~\ref{curves} suggests that there are
some deep minima of the similarity function even in cases when many word
correspondences are incorrect. An algorithm currently under construction
therefore searches for many local minima, and tries to find out what word
correspondences are the most reliable ones. In order to limit the search
space, translations that are known beforehand can be used as anchor points.

Future work will deal with the following as yet unresolved problems:
\begin{itemize}
\item{Computational limitations require the vocabularies to be limited to
subsets of all word
types in large corpora. With criteria like the corpus frequency of a word, its
specificity for a
given domain, and the salience of its co-occurrence patterns, it should be
possible to make
a selection of corresponding vocabularies in the two languages. If
morphological tools and
disambiguators are available, preliminary lemmatization of
the corpora would be desirable.}
\item{Ambiguities in word translations can be taken into account by working
with continuous
probabilities to judge whether a word translation is correct instead of making
a binary decision.
Thereby, different sizes of the two matrices could be allowed for.}
\end{itemize}
It can be expected that with such a method the quality of the results depends
on the thematic
comparability of the corpora, but not on their degree of parallelism. As a
further step, even
with non parallel corpora it should be possible to locate comparable passages
of text.

\section*{Acknowledgements}

I thank Susan Armstrong and Manfred Wettler for their support
of this project. Thanks also to Graham Russell and three
anonymous reviewers for valuable comments on the manuscript.

\end{document}